\documentclass{article}

\usepackage{arxiv}

\usepackage{times}  
\usepackage{helvet}  
\usepackage{courier}  
\usepackage[hyphens]{url}  
\usepackage{graphicx} 
\usepackage[numbers]{natbib}  
\usepackage{caption} 

\usepackage[utf8]{inputenc} 
\usepackage[T1]{fontenc}    
\usepackage{hyperref}       
\usepackage{url}            
\usepackage{booktabs}       
\usepackage{nicefrac}       
\usepackage{microtype}      
\usepackage{lipsum}
\usepackage{fancyhdr}       
\usepackage{graphicx}       
\graphicspath{{media/}}     
\usepackage{amsmath,amsfonts}
\usepackage{cleveref}       
\usepackage{algorithmic}
\usepackage{algorithm}
\usepackage{adjustbox}
\usepackage{array}
\usepackage{textcomp}
\usepackage{stfloats}
\usepackage{url}
\usepackage{verbatim}
\usepackage{paralist}
\usepackage{graphicx}
\usepackage{bbm}
\usepackage{multirow}
\usepackage{amssymb}
\usepackage{subcaption}
\usepackage{algorithm}
\usepackage{algorithmic}
\newtheorem{theorem}{Theorem}[section]

\newtheorem{lemma}[theorem]{Lemma}

\title{\emph{To Prune or not to Prune: }:\\ A Chaos-Causality Approach to Principled Pruning of Dense Neural Networks}

\author{
  Rajan Sahu\\
  Dept. of CSIS, Birla Institute of Technology and Sciences Pilani, Pilani\\
  \texttt{f20190572@pilani.bits-pilani.ac.in,}
\And
Shivam Chadda\\
BITS Pilani, Goa Campus\\
\texttt{f20190704@goa.bits-pilani.ac.in}
\And
Nithin Nagaraj\\
Consciousness Studies Programme, National Institute of Advanced Studies, Bangalore, India\\
\texttt{nithin@nias.res.in}
\And
Archana Mathur\\
Dept. of Information Science and Engineering, Nitte Meenakshi Institute of Technology, India\\
\texttt{archana.mathur@nmit.ac.in}
\And
Snehanshu Saha\\
Dept. of CSIS and APPCAIR,  BITS Pilani Goa, India and  HappyMonk AI\\
\texttt{snehanshus@goa.bits-pilani.ac.in}
}

\begin{document}
\maketitle

\begin{abstract}
Reducing the size of a neural network (pruning) by removing weights without impacting its performance is an important problem for resource constrained devices. In the past, pruning was typically accomplished by ranking or penalizing weights based on criteria like magnitude and removing low-ranked weights before retraining the remaining ones. Pruning strategies may also involve removing neurons from the network in order to achieve the desired reduction in network size. We formulate pruning as an optimization problem with the objective of minimizing misclassifications by selecting specific weights. To accomplish this, we have introduced the concept of chaos in learning (Lyapunov exponents) via weight updates and exploiting  causality to identify the causal weights responsible for misclassification. Such a pruned network maintains the original performance and retains feature explainability
\end{abstract}


\section{Introduction}

Designing a neural network architecture is a critical aspect of developing neural networks for various AI tasks, particularly in deep learning. One of the fundamental challenges in designing neural networks is finding the right balance between model complexity and sample size, which can have a significant impact on the network's performance. In general, a larger network with more parameters (overparameterized) can potentially learn more complex functions and patterns from the data \cite{10.1145/3133327}. However, 
challenge in network architecture design is to find the right balance between model complexity and sample size, so that the network can learn to generalize well to new, unseen data. In this context, 
over-parameterized networks \cite{zou2018stochastic}, \cite{adaswarm} have become increasingly popular in the deep learning era due to their ability to achieve high expressivity and potentially better generalization performance \cite{Shen2019NonlinearAV}. The idea behind over-parameterization is to increase the number of parameters in the network beyond what is strictly necessary to fit the training data and remarkable generalization to test data. However, such networks still require a number of assumptions and hyperparameter tuning to guarantee optimal performance.

Pruning techniques should reduce the number of parameters in a neural network without compromising its accuracy. 
However, it is important to ponder why we do not simply train a smaller network from scratch to make training more efficient. The reason is that the architectures obtained after pruning are typically more challenging to train from scratch \cite{Liu2018RethinkingTV}, and they often result in lower accuracy compared to the original networks. Therefore, while standard pruning techniques can effectively reduce the size and energy consumption of a network, it does not necessarily lead to a more efficient training process.


\subsection{Problem statement and Contributions}
We pose a broad \textit{Research Question} here: Is there a principled approach to pruning overparameterized, dense neural networks to a reasonably good sparse approximation such that performance is not compromised and explainability is retained? It is well known that dense neural network training and particularly weight updates via SGD have some element of chaos \cite{Zhang2021EdgeOC},\cite{Herrmann2022ChaoticDA}. We expect that, between the successive weight updates due to SGD and miss-classification, there is some observed causality and non-causal weights (parameters) can be pruned, leading to a sparse network i.e. some weight updates cause reduction in network (training) loss and some do not!  
Can we train a dense network till a few epochs to derive a pruned architecture for the derivative to run for the remaining epochs and produce performance metrics in the $\epsilon-$ball of the original, dense network? Does this sparse network also train well, verified with Shapley \cite{shap}  and WeightWatcher (\textit{WW}) \cite{Martin2020PredictingTI} tests.?
Specifically, we contribute to the following:
\begin{itemize}
  \item Present a unique and unifying framework on chaos and causality for deep network pruning. The unifying framework uses LE \cite{Kondo2021LyapunovEA} and Granger causality (GC) \cite{Granger1969InvestigatingCR} tandem.
    \item Propose novel pruning architectures, Lyapunov Exponent Granger Causality driven Fully Trained Network (LEGCNet-FT) and Lyapunov Exponent Granger Causality driven Partially Trained Network (LEGCNet-PT).
    \item LEGCNet-FT and LEGCNet-PT compare very well in performance and other baselines, Random \cite{random} and Magnitude based \cite{magnitude} pruning techniques.
    \item Establish feature consistency of LEGCNet-FT and LEGCNet-PT in explainability. 
    \item Verify empirically that the proposed architectures for pruning are not over-trained and obviously not overparameterized but are still able to generalize well, on diverse data sets while saving Flops (FLoating-point OPerationS). We accomplish this via the \textit{WW} test.
\end{itemize}
In summary, we propose pruning techniques, \textit{LEGCNet-FT and LEGCNet-PT} which perform at par with the dense, unpruned architecture and the existing pruning baselines. \textit{While maintaining consistent performance, these techniques also help reduce epochs to converge and Flops to compute while maintaining feature consistency with their dense counterparts and ensuring proper training across layers validated via \textit{WW} statistics.}

\section{Technical Motivation}

Chaos, causality and the manifestation of the Lottery Ticket Hypothesis are the key motivation behind our proposed pruning mechanism. 
\subsection{Gradient Descent and Low Dimensional Chaos} Is the process of updating weights in backpropagation via Gradient Descent chaotic? Is there an alternative interpretation of the minima in the weight landscape via low-dimensional chaos? The weight update in SGD is written as $w_{i+1}\leftarrow w_i-\eta_i\nabla_w f(w_i)$ may be thought of as a discretization to the first order ODE: $w'(t)=-\nabla_w f(w_i)$. The minimizer of the SGD is therefore conceived as a stable equilibrium of the ODE. That is, the minimum, $w^*$ can be thought of as a fixed point to the iterates $w_{i+1}=G(\eta_i,w_i) \equiv w^*=G(\eta_i,w^*)$. 

\noindent \textbf{Empirical evidence of chaos in back propagation:}
We performed a series of experiments on different datasets to check Sensitive Dependence On Initial Conditions (SDIC) for weight initialization, on single hidden layer neural network. The weight initialization matrix $W_{ij}$ followed Gaussian distribution $(W_{ij} \sim N(0, \sigma^2) )$. We recorded two set of executions - one with intial weight $w_{11}: w_{ij} \forall i \in {1..h}, \forall j \in {1..n}$, where $n, h$ are the number of input and hidden neurons - and another, with infinitesimal perturbation ($w_{11} + \delta$) keeping other parameters same. Each time, the network was trained via gradient descent and weight series were recorded. The method was repeated for the second weight connection $w_{ij}, i=1,j=2$ and for all the weights on the network. Later, the Lyapunov exponents were computed by using the TISEAN package \cite{Hegger1998PracticalIO} on the recorded weight series to measure the perturbed trajectory due to initial perturbation $\delta$. We observed positive Lyapunov exponents which marked the presence of some chaotic behavior in gradient descent.

\subsection{Chaos and Causality} 
One way to address the issue of explainability in AI/machine learning is to seek causal explanations for choices made in the learning process. Conversely, a learning process that incorporates choices made out of causal considerations is easier to explain and interpret. This is the motivation behind using causality based criteria for the choice of what to prune (or not prune) in this study. To this end, we employ Granger Causality (GC)~\cite{granger1969investigating}. 

The principle of pruning that is causally-informed is formulated as follows. Those connections (weights) in the learning network that do not causally impact the loss are chosen for pruning. To determine the causal impact of a particular connection to the loss, we perform GC between the windowed Lyapunov exponents (LE) of the weight time series for that connection and the classification accuracy. The rationale behind this is the intuition that the chaotic signature of weight updates inform learning. A biological inspiration for chaotic signatures as a marker for learning is the empirical fact that neurons in the human brain exhibit chaos~\cite{faure2001there, korn2003there} at all spatio-temporal scales. Starting from single neurons to coupled neurons to network of neurons to different areas of brain -- chaos has been found to be ubiquitous to brain~\cite{korn2003there}. Chaotic systems are known to exhibit a wide range of patterns (periodic, quasi-periodic and non-periodic behaviors), are very robust to noise and enable efficient information transmission~\cite{nagaraj2009multiplexing}, processing/computation~\cite{ditto2015exploiting, kuo2005chaos} and classification~\cite{balakrishnan2019chaosnet}. There is also some evidence to suggest that weak chaos is likely to aid learning~\cite{sprott2013chaos}.  Thus our choice of testing causal strength between LE (a value $>0$ is a marker of chaos) and classification accuracy as a criterion for pruning to yield sparse subnetworks that capture the learning of the task at hand.

\subsection{Lottery ticket hypothesis} The "lottery ticket hypothesis" \cite{Frankle2018TheLT} is a concept in neural network pruning that suggests that within a dense and over-parameterized neural network, there exist sparse sub-networks that can be trained to perform just as well as the original dense network. Any fully-connected feed-forward network $f^d(x;\phi)$, with initial parameters $\phi$ when trained on a training set D, $f^d$ achieves a test accuracy $a$ and error $e$ at iteration $j$. 
Our work LEGCNet, validates the lottery ticket hypothesis by finding the "winning-ticket", $m$, to construct the sparse network, $f^s$ such that ${acc^s} \geq  {a}$  and ${j^s} < {j}$ where $\left \| \phi \right \|  \gg   \left \| m \right \| $.

The remainder of the paper is organized to present the key methodologies used to develop the pruning technique (section 3), followed by a detailed experimental setup (section 4) and strong empirical evidence of the proposed technique in contrast to the baselines (section 5,6).

\section{Tools and Methodology}

\subsection{Definitions:}
\begin{itemize}
    \item Dense Neural network - Let $f^d$ be a dense neural network of depth $l$ and width $h$ defined as 
\begin{equation} \label{eq:n1}
    f^d(x) = W^d_i \sigma _i( W^d_{i-1} \sigma _i(...W^d_1(x)))
\end{equation}
where $W^d_i$ is the weight matrix for layer $i$ such that $i \in {1..l}$.  
    \item Sparse Neural Network - Let $f^s$ be sparse neural network of the same architecture as $f^d$, with depth $l$ and width $h$. 
    \item Two approaches: LEGCNet-FT and LEGCNet-PT - In order to validate the working of LEGCNet, we divided the method into two discrete approaches. In one approach, the entire training weight-series is used for computing Lyapunov exponents, testing Granger causality and for the identification of causal weights. Essentially, the dense network is trained till convergence and the approach is called LEGCNet-Full Train (LEGCNet-FT). In the second approach, LEGCNet-PT, the network is trained only till certain epochs (10\% of the total iterates) and these few weight-updates are captured for identifying the causal weights.    
\end{itemize}

\subsection{WeightWatcher (\textit{WW}) as a diagnostic tool} \textit{WW} is a powerful open-source diagnostic tool designed for analyzing Deep Neural Networks (DNN). 
\textit{WW} analyzes each layer by plotting the empirical spectral distribution (ESD), which represents the histogram of eigenvalues from the layer's correlation matrix. Additionally, it fits the tail of the ESD to a (truncated) power law and presents these fitted distributions on separate axes. This visualization approach provides a clear representation of the eigenvalue distribution and highlights the presence of heavy-tailed behavior in the network's layers. In general, the ESDs observed in the best layers of high-performing DNNs can often be effectively modeled using a Power Law (PL) function. The PL exponents, denoted as alpha, tend to be closer to 2.0 in these cases indicating a heavy-tailed behavior in the layer's correlation matrix. 
\subsection{SHAP as a diagnostic tool}
SHAP (SHapley Additive exPlanations) is a game-theoretic technique utilized to provide explanations for the output of machine learning models. 
SHAP enables a comprehensive understanding of the contributions made by different features in the model's output, facilitating insightful explanations for its decision-making process. If a network is pruned according to some underlying principles, then the consistency in feature explainability is maintained before and after pruning i.e. the features which explain the outcome before pruning (fully connected, dense network) remain consistent on the pruned network.
\subsection{Windowed Weight Updates:} 
Consider a neural network with $n$ inputs and $r$ outputs, $l$ hidden layers of $h$ neurons, and, the input vector denoted as $x \in  R^n$, 
The network when trained by SGD generates a sequence of weight updates represented by 
$w^{ji} = \left [ w_{0}^{ji}, w_{1}^{ji}, ..., w_{k}^{ji}, ..., w_{K}^{ji} \right ]
$
where $w_{k}^{ji}$ is the weight of ith neuron of the input layer and jth neuron of the hidden layer at kth iteration. Considering, the weights being collected for initial few epochs, the weight iterates for hidden layer and output layer are 
$
    W_h=\left\{ w^{ji} \right\}, W_o=\left\{ w^{kj} \right\} \, \, \, \, \, \forall i \in \left\{ 1,..,n \right\},  \forall j \in \left\{ 1,..,m \right\}, \forall k \in \left\{ 1,..,r \right\}
    \label{eq:3}
$ 
An infinitesimal perturbation $\delta_0$ is introduced in the initial weight $w^{11}$, given as  $w^{11\delta_0} = w^{11}+ \delta_0$, keeping other parameters - weights (initialization), learning rate, optimizer, and loss function- same. The network is then retrained with the perturbed weight, and the weights updates are recorded again as 
$
    W_h^{\delta_0}=\left\{ w^{ji \delta_0} \right\}, W_o^{\delta_0}=\left\{ w^{kj \delta_0} \right\} \, \, \, \, \, \, \, \, \, \,  \forall i \in \left\{ 1,..,n \right\},  \forall j \in \left\{ 1,..,m \right\}, \forall k \in \left\{ 1,..,r \right\}
    \label{eq:2}
    $    
A difference series obtained by subtracting perturbed weights from initial weights is
$
    \delta W_h=\left\{\delta w^{ji} \right\}, \delta W_o=\left\{ \delta w^{kj} \right\} \, \, \, \, \, \, \, \, \, \,  \forall i \in \left\{ 1,..,n \right\},  \forall j \in \left\{ 1,..,m \right\}, \forall k \in \left\{ 1,..,r \right\}
    \label{eq:1}
$.
We divide the weight series $\delta w^{ji}$ into $K$ windows, $w^{ji} = \bigcup_{l=1}^{K} w^{ji\left ( l \right )}$, and compute the Lyapunov exponent of all the windowed-weight trajectories. 
The series of the Lyapunov exponent $\lambda$ of the windowed-weight trajectories $w^{ji(K)}$  are represented using the notation $\left \{ \lambda^{ji\left \{ 1 \right \}},\lambda^{ji\left \{ 2 \right \}},...,\lambda^{ji\left \{ K \right \}}  \ \right \}$. Additionally, we record the accuracy at every window during training $\forall l \in 1, \ldots, K$, captured for weights at $w^{ji(l)}$ and $w^{kj(l)} \forall i \in \left\{ 1,..,n \right\},  \forall j \in \left\{ 1,..,m \right\}, \forall k \in \left\{ 1,..,r \right\}$.

\begin{figure*}[t]
    \centering    	\includegraphics[width=0.65\textwidth]{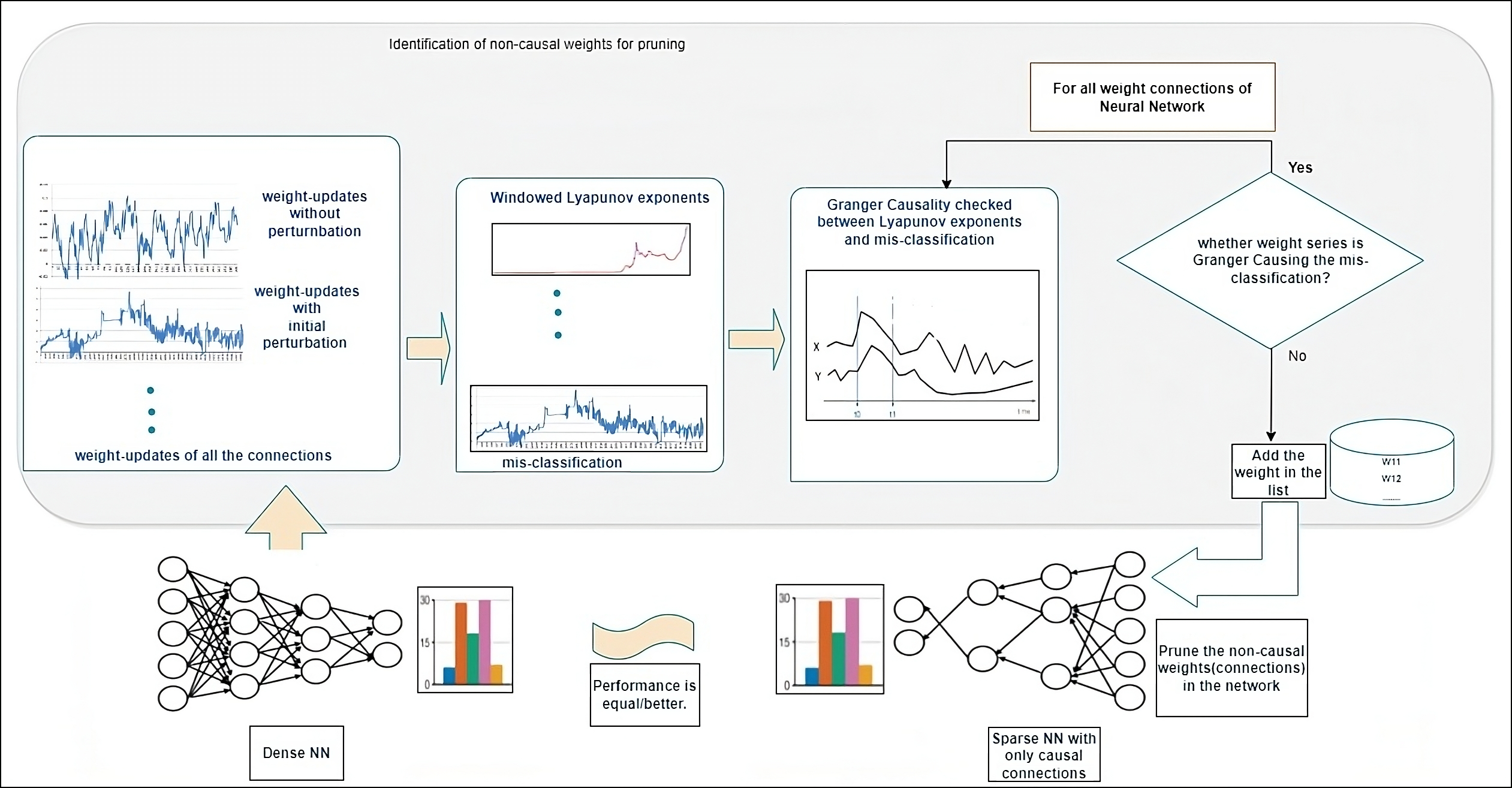}
    \caption{Weights pruned via LEGCNet method for selecting connections in sparse neural network}
    \label{fig:methodology}
\end{figure*}
\subsection{Approximation capability of LEGCNet}
The main theorem and lemma in this section build that the LEGCNet pruned network ($f^s$) is $\epsilon$-close to $f^d$ with probability 1-$\delta$ \cite{Qian2021APA}. Consider $f^d$ to be a dense neural network defined as 
\begin{equation} \label{eq:n}
    f^d(x) = W^d_i \sigma _i( W^d_{i-1} \sigma _i(...W^d_1(x)))
\end{equation}
where $W^d_i$ is the weight matrix for layer $i$ such that $i \in {1..l}$ and $h_i > h_0, h_i > h_l, \forall i \in 1...(l-1)$. We assume that for network in (\ref{eq:n}), $\sigma_i$ is $L_i$-Lipschitz and weight matrix $W^d_i$ is initialized from uniform distribution $\mathit{U}[\frac{-K}{\sqrt{max(h_i,h_{i-1})}},\frac{K}{\sqrt{max(h_i,h_{i-1})}}]$, for some constant $K$.
\begin{theorem} 
(Approximation Capability of the Sparse Network) Let $\epsilon >0$, $\delta>0$, $\alpha \in (0,1) $ such that the, for some constants $K_1,K_2,K_3, K_4, K_5$, 
\begin{equation*}
    h \geqslant max \left \{ K_1^\frac{1}{\alpha}, \left ( \frac{K_2}{\epsilon} \right )^\frac{1}{\alpha}, \left ( \frac{K_3}{\delta } \right )^\frac{1}{\alpha},K_4 + K_5 log\left ( \frac{1}{\delta } \right )   \right \}
\end{equation*}
then sparse network $f^s$ obtained from LEGCNet by the mask $m$, and pruning the weights $W^d_i$,  $\forall i \in {1...l}$ is $\epsilon$-close to $f^d$, with the probability ($1- \delta$), i.e.
\begin{equation*}
    \underset{x \in B_{d0}}{\textup{sup}} \left \| f^s(x) -f^d(x) \right \|_2 \leq \epsilon
\end{equation*}
\end{theorem}
\noindent \textbf{Remark:} Lipschitz property of the activation functions \cite{Saha2020EvolutionON} is a necessary condition to validate the approximation capability of the proposed sparse network. We have used \textit{sigmoid activation} in the sparsely trained/pruned network.
\begin{lemma}
Sigmoid activation is Lipschitz.
\end{lemma}
\noindent \textbf{Proof:} If a function $f(x)$ is Lipschitz continuous, then: $ \left \| f(x) - f(y) \right \| \leq K \left \| x-y \right \| \equiv \left \| f'(x) \right \| \leq K$. If $K<1, f$ is a contraction map as well. We know that Sigmoid, $\sigma (x) = \frac{1}{1+e^{-x}};$ and $ \left \| {\sigma}' \right \| = \left \| \sigma (x) (1-\sigma (x) )  \right \|$. It's easy to follow that : $ \left \| \sigma (x) (1-\sigma (x) )  \right \| \leq \left \| \sigma (x)  \right \| \left \| 1-\sigma (x)  \right \| 
    \leq C_1, C_2$. Since 0$ \leq C_1$, $C_2 \leq$ 1,   $C_1*C_2 =\delta \leq 1$. Hence sigmoid is Lipschitz. 

\subsection{Methodology Overview (refer figure \ref{fig:methodology})}
Our pruning method was developed as follows. Initially, we trained a simple Multi-layer Perceptron (MLP) on a given dataset. Throughout the training process, we recorded the weights at each iteration, resulting in a time series of weight values. These weight time series were subsequently utilized to estimate the Lyapunov exponents, a measure of chaotic behavior, using the TISEAN package in conjunction with MATLAB scripts. This estimation was performed using a sliding window approach, generating a time series of Lyapunov exponents.

Based on our experiments, the time series of Lyapunov exponents consistently exhibited positive values, indicating the presence of chaotic elements in the weight time series. Consequently, our study focused on understanding whether these weights Granger caused the model's accuracy. Any weights that did not demonstrate this causal relationship were pruned before conducting subsequent model runs (code is uploaded in supplementary, zipped file).


\section{Experimental set-up}
In our study, we employed Python3.10 and Matlab R2022a to conduct experiments on a single hidden layer neural network on various datasets. 
Our experiments were conducted on Ryzen 9 3900XT Desktop Processor with 32GB RAM and 1TB HDD. 
During training, we stored the weight updates for every connection in CSV files. We assumed a window size of 200 iterates and computed the Lyapunov exponent for each weight connection on every window. Further, we calculated the training and test accuracy on every window, to capture the misclassification rates. Thus, we obtained a sequence of windowed-Lyapunov exponents and windowed accuracies for every connection. We then computed the Granger causality between the windowed Lyapunov exponents and the misclassification rate to identify weight connections that Granger caused misclassification. In the process of pruning the network, we removed the connections for which the Lyapunov exponents were found to Granger cause misclassification. After pruning, we reran the experiment by keeping the initial weights, optimizers, and other hyperparameters the same as unpruned network. We extended the work on different datasets and recorded the epochs and accuracies of the pruned network.  
Interestingly, we observed that the accuracies of the sparse network exceeded those of the dense network.
\par In our study, we conducted experiments in two parts. In the first part, we trained the neural network until convergence (LEGCNet-FT) and computed windowed-Lyapunov exponents for each weight connection as well as windowed accuracies. We then computed the Granger causality and pruned the network by removing connections that were found to cause misclassification. In the second part of the experiment, we trained the network only for few epochs (LEGCNet-PT) and used these initial weight updates to compute windowed-Lyapunov exponents and accuracies, repeating the same procedure as in the first part of the experiment. Finally, we compared the performance of the pruned network (LEGCNet-FT and LEGCNet-PT) to that of the original network. The results of all these experiments were recorded and presented in tables. Code is available as supplementary file.





\begin{table*}[]
\centering
\scalebox{0.65}{
\begin{tabular}{|l|r|r|r|r|r|r|r|r|r|r|}
\hline
\multicolumn{1}{|c|}{Dataset (hidden neurons)} & \multicolumn{1}{c|}{\begin{tabular}[c]{@{}c@{}}Flops -\\ DN\end{tabular}} & \multicolumn{1}{c|}{\begin{tabular}[c]{@{}c@{}}Flops -\\ LEGCNet-FT\end{tabular}} & \multicolumn{1}{c|}{\begin{tabular}[c]{@{}c@{}}Non causal \\ Weights\end{tabular}} & \multicolumn{1}{c|}{\begin{tabular}[c]{@{}c@{}}Epochs\\ DN\end{tabular}} & \multicolumn{1}{c|}{\begin{tabular}[c]{@{}c@{}}Epochs \\ LEGCNet-FT\end{tabular}} & \multicolumn{1}{c|}{\begin{tabular}[c]{@{}c@{}}Accuracy \\ DN\end{tabular}} & \multicolumn{1}{c|}{\begin{tabular}[c]{@{}c@{}}Accuracy \\ LEGCNet-FT\end{tabular}} & \multicolumn{1}{c|}{\begin{tabular}[c]{@{}c@{}}F1-score\\ DN\end{tabular}} & \multicolumn{1}{c|}{\begin{tabular}[c]{@{}c@{}}F1-score\\ LEGCNet-FT\end{tabular}} & \multicolumn{1}{c|}{\begin{tabular}[c]{@{}c@{}}\%Pruned \\ LEGCNet-FT\end{tabular}} \\ \hline
cancer(6)  & 60 & 54 & 6 & 70 & 38 & 0.8759 & 0.8686 & 0.8656 & 0.8570 & 10 \\ \hline
Titanic (8) & 56 & 52 & 4 & 43 & 20 & 0.7225 & 0.7177 & 0.6948 & 0.6843 & 7.14 \\ \hline
Banknote (8) & 40 & 36 & 4 & 9 & 7 & 0.9018 & 0.8909 & 0.9016 & 0.8906 & 10 \\ \hline
Iris (6) & 42 & 40 & 2 & 182 & 166 & 0.9 & 0.9 & 0.9124 & 0.9419 & 4.76 \\ \hline
Iris(3 features)(6) & 36 & 26 & 10 & 139 & 126 & 0.9 & 0.9 & 0.933 & 0.904 & 27.78 \\ \hline
Vowel (4) & 36 & 34 & 2 & 36 & 14 & 0.7462 & 0.7538 & 0.7307 & 0.74 & 5.56 \\ \hline
Mnist (50, 30)  & 41000 & 40861 & 139 & 27 & 30 & 0.9121 & 0.9165 & 0.8669 & 0.8781 & 0.34 \\ \hline
\end{tabular}}
\caption{Comparing the performance of sparse network with dense networks by using LEGCNet across different datasets. The approach used for finding non-causal weights is LEGCNet-Fully Trained. Accuracies used are on the Test set. Dense Network - DN, Sparse Network - SN}
\label{tab:LEGCNet-FT}
\end{table*}

\begin{table*}[]
\centering
\scalebox{0.65}{
\begin{tabular}{|l|r|r|r|r|r|r|r|r|r|r|}
\hline
\multicolumn{1}{|c|}{Data($Epochs^*$)} & \multicolumn{1}{c|}{\begin{tabular}[c]{@{}c@{}}Flops (SN) -\\ LEGCNet-FT\end{tabular}} & \multicolumn{1}{c|}{\begin{tabular}[c]{@{}c@{}}Flops (SN) -\\ LEGCNet-PT\end{tabular}} &  \multicolumn{1}{c|}{\begin{tabular}[c]{@{}c@{}}Epochs (SN)\\ LEGCNet-FT\end{tabular}} & \multicolumn{1}{c|}{\begin{tabular}[c]{@{}c@{}}Epochs (SN)\\ LEGCNet-PT\end{tabular}} & \multicolumn{1}{c|}{\begin{tabular}[c]{@{}c@{}}Accuracy (SN)\\ LEGCNet-FT\end{tabular}} & \multicolumn{1}{c|}{\begin{tabular}[c]{@{}c@{}}Accuracy (SN)\\ LEGCNet-PT\end{tabular}} & \multicolumn{1}{c|}{\begin{tabular}[c]{@{}c@{}}F1-score (SN)\\ LEGCNet-FT\end{tabular}} & \multicolumn{1}{c|}{\begin{tabular}[c]{@{}c@{}}F1-score (SN)\\ LEGCNet-PT\end{tabular}} & \multicolumn{1}{c|}{\begin{tabular}[c]{@{}c@{}}\%Pruned \\ LEGCNet-PT\end{tabular}} \\ \hline
Cancer(12) & 54 & 42  & 38 & 24 & 0.8686 & 0.8759 & 0.8570 & 0.8643 & 30 \\ \hline
Titanic(3) & 52 & 29 & 20 & 36 & 0.7177 & 0.7249 & 0.6843 & 0.6969 & 48.21 \\ \hline
Banknote(2) & 36 & 37 & 7 & 6 & 0.8909 & 0.8945 & 0.8906 & 0.8943 & 7.5 \\ \hline
Iris(20) & 40 & 40 & 166 & 173 & 0.9 & 0.9000 & 0.9419 & 0.9124 & 4.76 \\ \hline
Iris 3f(20)& 26 & 28  & 126 & 135 & 0.9 & 0.9333 & 0.9040 & 0.9330 & 22.22 \\ \hline
Vowel(5) & 34 & 28 & 14 & 20 & 0.7538 & 0.7538 & 0.7400 & 0.7441 & 22.22 \\ \hline
MNIST(5) & 40861 & 40867  & 30 & 29 & 0.9165 & 0.9152 & 0.8781 & 0.8408 & 0.32 \\ \hline
\end{tabular}}
\caption{LEGCNet-PT: Comparing Sparse Networks where causality is computed from two different training scenarios - when the network is trained fully (LEGCNet-FT), and when trained till a few epochs (LEGCNet-PT); Iris 3f indicates IRIS dataset with 3 features; $Epochs^*$ - No of epochs used for computing non-causal weights in LEGCNet-PT }
\label{tab:LEGCNet-PT}
\end{table*}

\section{Performance comparison - dense network, LEGCNet-FT, and LEGCNet-PT} We ran the experiments on seven tabular datasets - Cancer, Titanic, Banknote, Iris, Iris (3 features) Vowel and MNIST. The datasets were divided into 80:20 train-test split and the code was run 5 times, each time maintaining different network initialization. The best results from every initialization are reported in tables \ref{tab:LEGCNet-FT} and \ref{tab:LEGCNet-PT}. We compared the Flops, \% pruned (fraction of parameters removed *100), accuracy, f1-scores, and epochs for all methods (dense, LEGCNet-FT and LEGCNet-PT). Table \ref{tab:LEGCNet-FT} shows the performance comparison of dense network and LEGCNet-FT. Remarkably, LEGCNet-FT achieves notable reductions in Flops without compromising accuracy. Furthermore, LEGCNet-FT converges significantly faster consuming a few epochs compared to the dense network. Specifically, for the Titanic, Vowel, and Cancer datasets, LEGCNet-FT achieves convergence in just half the number of epochs required by the dense network. Nonetheless, both network achieves a similar level of performance - accuracy, and f1-score- without significant differences, thus validating the lottery ticket hypothesis. Table \ref{tab:LEGCNet-PT} demonstrates the performance of LEGCNet-PT. It shows that LEGCNet-PT performs at par with its counterpart, in terms of convergence speed, Flops, accuracy and F1 scores.

\begin{table}[]
\centering
\scalebox{0.6}{
\begin{tabular}{|l|r|r|r|r|r|}
\hline
\multicolumn{1}{|c|}{Data} & \multicolumn{1}{c|}{\begin{tabular}[c]{@{}c@{}}Epochs (SN)\\ Random\end{tabular}} & \multicolumn{1}{c|}{\begin{tabular}[c]{@{}c@{}}Accuracy (SN)\\ Random\end{tabular}} & \multicolumn{1}{c|}{\begin{tabular}[c]{@{}c@{}}Accuracy (SN)\\ Magnitude\end{tabular}} & \multicolumn{1}{c|}{\begin{tabular}[c]{@{}c@{}}F1-score (SN)\\ Random\end{tabular}} & \multicolumn{1}{c|}{\begin{tabular}[c]{@{}c@{}}F1-score (SN)\\ Magnitude\end{tabular}} \\ \hline
Cancer & 42 & 0.8759 & 0.8905 & 0.8656 & 0.8814 \\ \hline
Titanic & 35 & 0.7201 & 0.7201 & 0.6842 & 0.6906 \\ \hline
Banknote & 6 & 0.8945 & 0.9018 & 0.8943 & 0.9015 \\ \hline
iris & 179 & 0.9000 & 0.9000 & 0.9124 & 0.9124 \\ \hline
Iris(3 features) & 160 & 0.9000 & 0.9333 & 0.9330 & 0.9330 \\ \hline
Vowel & 28 & 0.7462 & 0.7462 & 0.7229 & 0.7307 \\ \hline
Mnist & 31 & 0.9119 & 0.9121 & 0.8627 & 0.8669 \\ \hline
\end{tabular}}
\caption{ Results for Random and Magnitude-based pruning; }
\end{table}




\subsection {Diagnostics}
\begin{figure*}[]
\centering
\includegraphics[width=.28\textwidth]{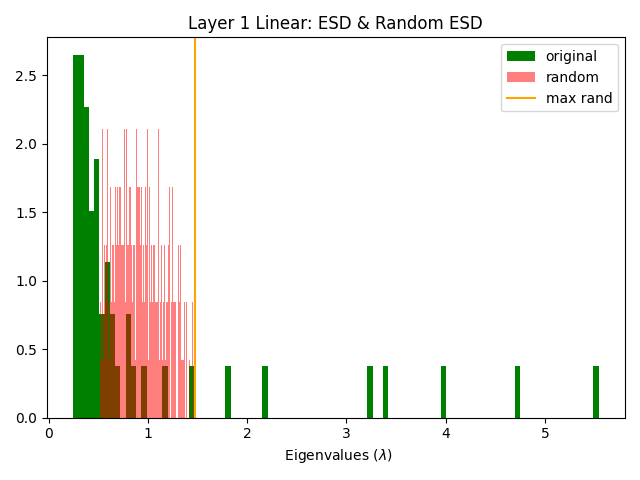}
\includegraphics[width=.28\textwidth]{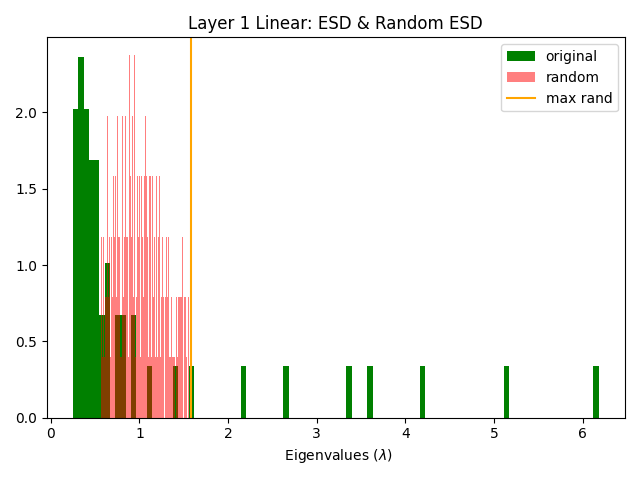}
\includegraphics[width=.28\textwidth]{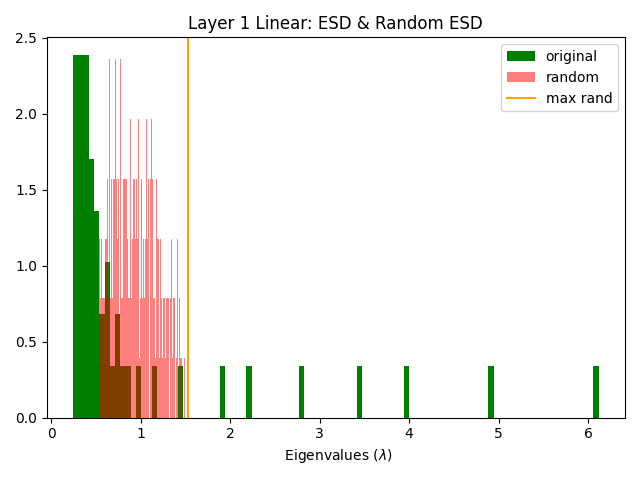}
\caption{\textbf{\textit{WW} plots for dense and LEGCNet-FT, and LEGCNet-PT networks on MNIST data (layer 1)}. Layers 2 and 3 are kept in the supplementary file (section A). Plots reveal the correct training of the proposed architectures. WW plots of MNIST for random  and magnitude pruning are in supplementary file section B.}
\label{fig:WW}
\end{figure*}

It is also crucial to examine the impact of our pruning technique on the training process of the model and determine if the relevance of feature importance is maintained and if the pruned network is properly trained or not. To accomplish this goal, we utilized two diagnostic tools, \textit{namely \textit{WW} and SHAP}. The \textit{WW} validates the compliance of our architecture, as reflected in the table \ref{tab:my-table} and figure \ref{fig:WW} plotted for MNIST. The alpha lies between 2.0 and 6.0 on every layer (table \ref{tab:my-table}). 
The ESD plots of the three types of training (dense, LEGCNet-PT, LEGCNet-FT) manifest a heavy-tailed distribution of eigenvalues on each layer indicating the layers are well-trained (figure \ref{fig:WW}). A careful observation at figure \ref{fig:WW} reveals the following: ESD plot of a layer, where the orange spike on the far right is the tell-tale clue; it's called a Correlation Trap \cite{LeCun1990SecondOP}. A Correlation Trap refers to a situation where the empirical spectral distributions (ESDs) of the actual (green) and random (red) distributions appear remarkably similar, with the exception of a small correlation shelf located just to the right of 0. In the random ESD (red), the largest eigenvalue (orange) is noticeably positioned further to the right and is separated from the majority of the ESD's bulk. This phenomenon indicates the presence of strong correlations in the layer, which can potentially affect the overall behavior and performance of the network. This is the case of a well-trained layer. 

\begin{figure*}[]
\centering
\includegraphics[width=.28\textwidth]{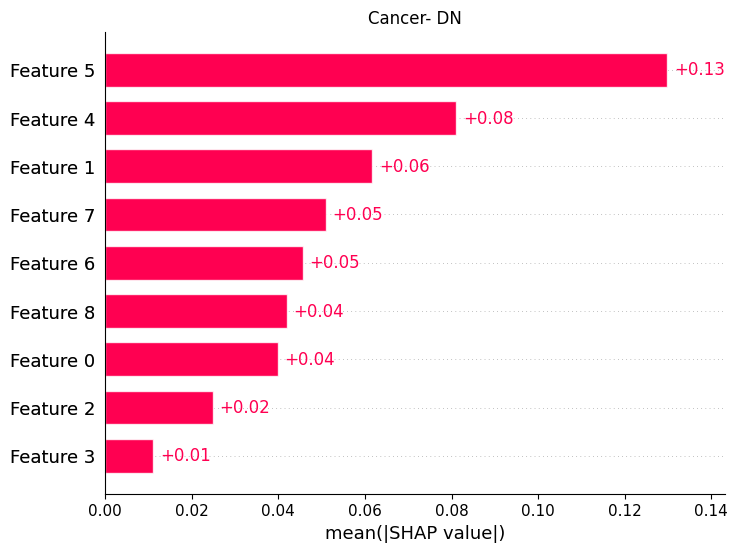}
\includegraphics[width=.28\textwidth]{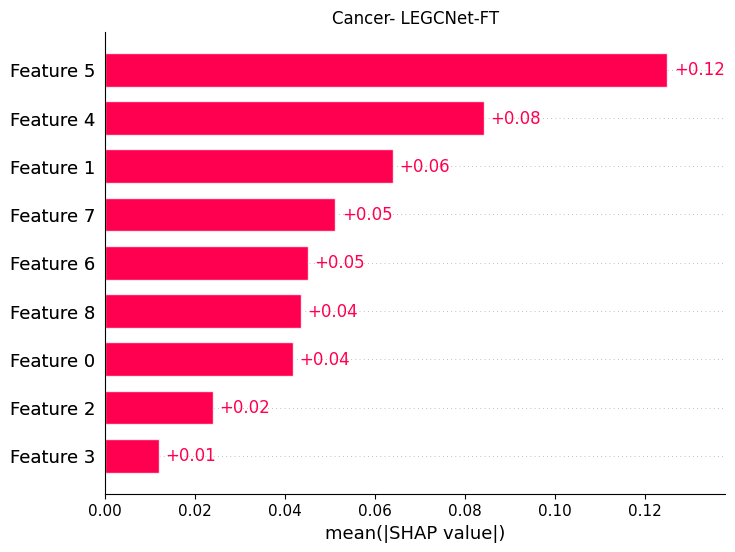}
\includegraphics[width=.28\textwidth]{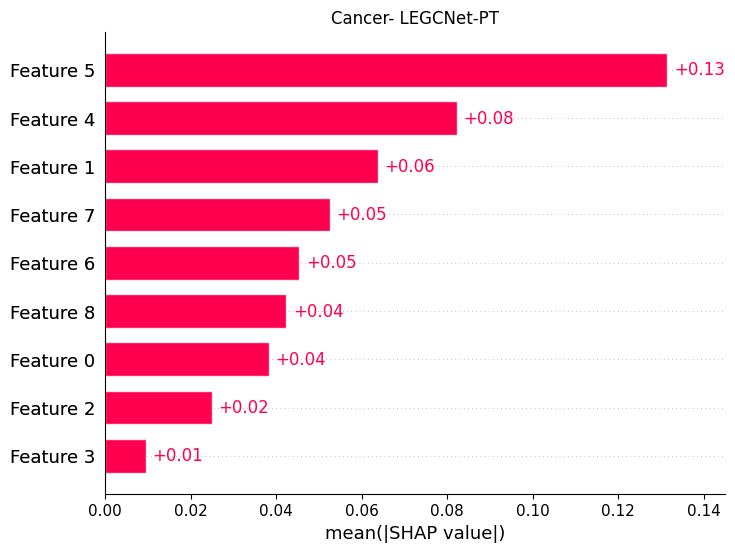}
\caption{Shap values and feature importance computed on Cancer dataset (Banknote and Titanic plots can be seen in the supplementary file section D)  for all the three models - Dense Network, LEGCNet-FT and LEGCNet-PT; the feature importance for the dense network is same as LEGCNet-FT and LEGCNet-PT }
\label{fig:shap}
\end{figure*}

\textit{SHAP:}
The SHAP values computed for the proposed pruning architecture as well as for the baselines - random and magnitude - indicate that the feature importance is maintained in LEGCNet-PT and LEGCNet-FT when compared with dense (figure \ref{fig:shap}). However, the baseline pruning methods (random, magnitude pruning) could not maintain the feature consistency as seen in the SHAP plots (supplementary file, section C). Though magnitude pruning shows feature consistency for Cancer, banknote, and Titanic datasets, the random pruning could not. 

\begin{table}[h]
\centering
\scalebox{0.70}{
\begin{tabular}{|l|rr|rr|rr|}
\hline
\multicolumn{1}{|c|}{} & \multicolumn{2}{c|}{Layer1: 784-50} & \multicolumn{2}{c|}{Layer2: 50-30} & \multicolumn{2}{c|}{Layer3: 30-10} \\ \hline
Model & \multicolumn{1}{c|}{alpha} & \multicolumn{1}{c|}{alpha\_w} & \multicolumn{1}{c|}{alpha} & \multicolumn{1}{c|}{alpha\_w} & \multicolumn{1}{c|}{alpha} & \multicolumn{1}{c|}{alpha\_w} \\ \hline
Dense & \multicolumn{1}{r|}{2.19} & 1.63 & \multicolumn{1}{r|}{1.51} & 1.88 & \multicolumn{1}{r|}{1.94} & 2.76 \\ \hline
LEGCNet-FT & \multicolumn{1}{r|}{2.24} & 1.64 & \multicolumn{1}{r|}{1.51} & 1.83 & \multicolumn{1}{r|}{2.29} & 3.20 \\ \hline
LEGCNet-PT & \multicolumn{1}{r|}{2.19} & 1.71 & \multicolumn{1}{r|}{1.51} & 1.85 & \multicolumn{1}{r|}{2.73} & 3.87 \\ \hline
Random & \multicolumn{1}{r|}{2.30} & 1.53 & \multicolumn{1}{r|}{1.70} & 1.95 & \multicolumn{1}{r|}{2.00} & 2.84 \\ \hline
Magnitude & \multicolumn{1}{r|}{2.19} & 1.63 & \multicolumn{1}{r|}{1.55} & 1.85 & \multicolumn{1}{r|}{1.96} & 2.78 \\ \hline
\end{tabular}}
\caption{Weight Watcher summary retrieved for all models trained on the same initialization for MNIST}
\label{tab:my-table}
\end{table}

\section{Discussion and Conclusion}

Unlike the current baselines, the percentage of pruned weights is significantly less. This is because only the non-causal weights are pruned, weights that play no role in impacting the loss/accuracy. 
We argue that the proposed strategy is efficient and accurate, with the additional benefit of passing network fitting and explainability tests, in addition to satisfying the lottery ticket hypothesis (tables \ref{tab:LEGCNet-FT} and \ref{tab:LEGCNet-PT} Experimental section.). One of the other salient features of LEGCNet is that, unlike other pruning methods, it does not need the dense network to be trained for the entire cycle of epochs to identify pruning candidates. Rather, such candidates are detected after a few initial epochs so that the retraining can start immediately. This reflects in reduced Flops without compromising key performance indicators (tables \ref{tab:LEGCNet-FT}, \ref{tab:LEGCNet-PT}). Notably, our architecture is validated for correct training via \textit{WW} Statistics as detailed in the diagnostic section previously.
The experimental results demonstrated that the proposed pruning method exhibits notable advantages in maintaining feature importance compared to the traditional random and magnitude pruning methods. The feature consistency remained relatively stable after employing the proposed pruning technique, which was not the case for the other two methods. 
In the case of random and magnitude pruning, significant fluctuations in feature importance were evident after pruning. These fluctuations could potentially hinder the interpretability of the underlying model. However, our proposed pruning method demonstrated remarkable resilience in preserving feature importance, with minimal perturbations observed in SHAP plot patterns enabling a more interpretable and transparent pruned model. 
For mission-critical tasks on edge devices such as predicting power consumption of applications \cite{alavani2023program} or forecasting real-time blood glucose prediction, feature explainability on pruned networks are critical as they help determine accurate prediction when dimensionality is a curse. 
It is crucial to emphasize that our primary focus in this investigation was on the effects of pruning, rather than achieving perfect classification performance. We compared three distinct pruning techniques: random pruning, magnitude pruning, and our pruning approach. Overall, the experimental outcomes validate the superiority of the proposed pruning method. 
The findings hold great promise for further advancements in network optimization and model explainability. 
Our pruning approach is yet to be tested on baseline architectures (Resnet, Densenet), and Large Language Models and  savings in carbon emission needs to be computed.
\section*{Acknowledgments}
Archana Mathur and Nithin Nagaraj would like to thank SERB-TARE (TAR/2021/000206) for supporting the work. Snehanshu Saha would like to thank the DBT-Builder project, Govt. of India (BT/INF/22/SP42543/2021) and SERB-SURE, DST.

\bibliography{main}
\bibliographystyle{plain}

\newpage
\appendix
\onecolumn

\section*{Appendix}
Detailed implementation is available at the Github repository: https://github.com/RAJAN13-blip/chaotic-pruning
\section{Plots on WeightWatchers run on different datasets} 
\begin{figure*}[h]
\centering
\includegraphics[width=.33\textwidth]{media/dense/ww.layer1.randesd1.png}
\includegraphics[width=.33\textwidth]{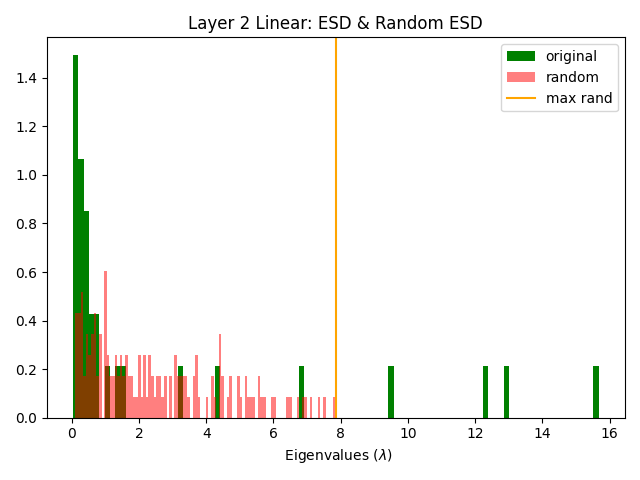}
\includegraphics[width=.33\textwidth]{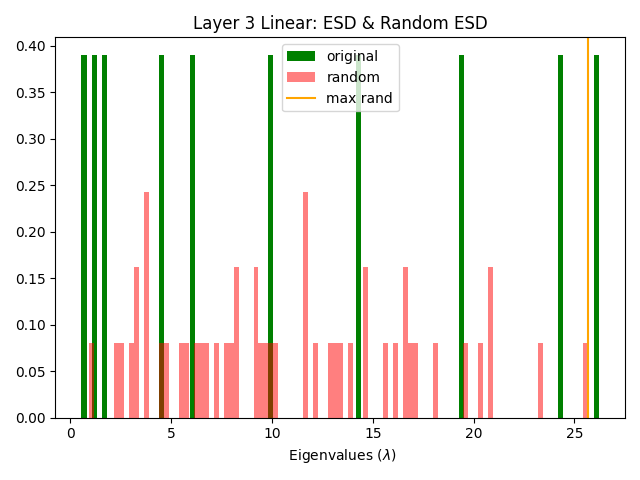}
\includegraphics[width=.33\textwidth]{media/FT/ww.layer1.randesd1.png}
\includegraphics[width=.33\textwidth]{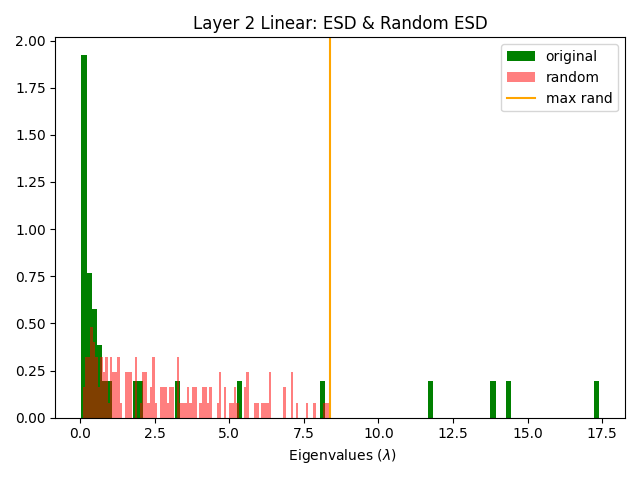}
\includegraphics[width=.33\textwidth]{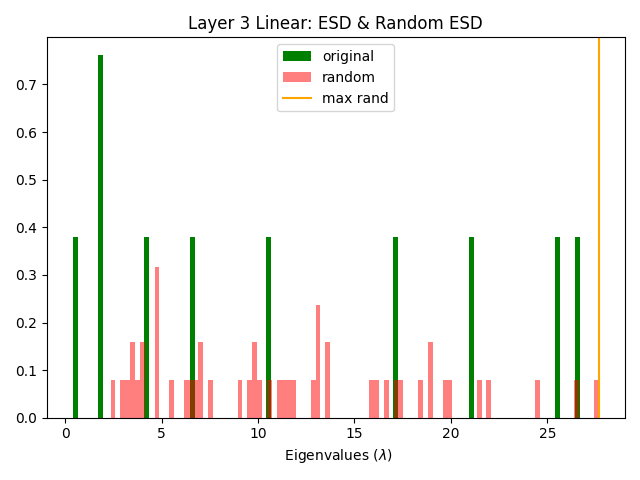}
\includegraphics[width=.33\textwidth]{media/PT/ww.layer1.randesd1.png}
\includegraphics[width=.33\textwidth]{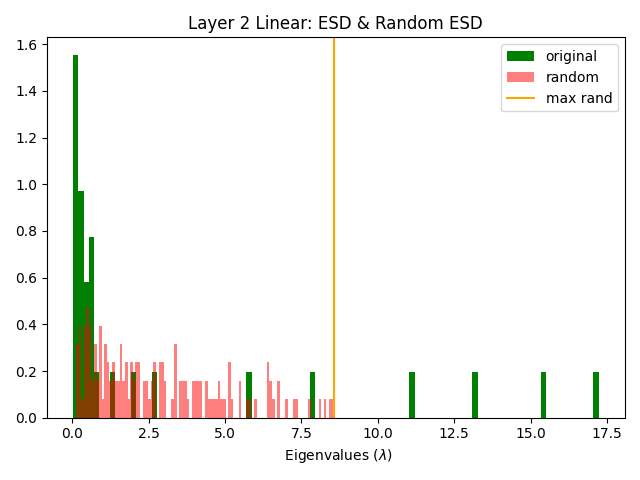}
\includegraphics[width=.33\textwidth]{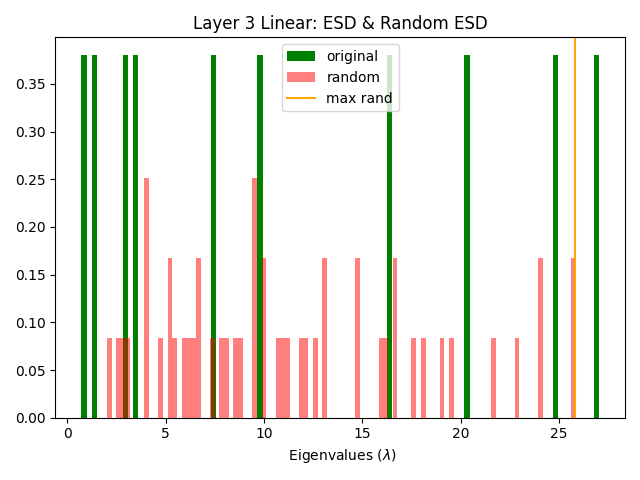}
\caption{\textbf{WW plots for layer-wise Dense and LEGCNet-FT (figures 3 and 4) and LEGCNet-PT (figures 5 and 6) networks on MNIST data}. Plots reveal the correct training of the proposed architectures.}
\label{fig:WW1}
\end{figure*}

It is also crucial to examine the impact of our pruning technique on the training process of the model and determine if the relevance of feature importance is maintained and if the pruned network is properly trained or not. To accomplish this goal, we utilized two diagnostic tools, \textit{namely WeightWatcher (WW) and SHAP}.  The alpha lies between 2.0 and 6.0 on every layer, however, layer 3 of the dense network and the (baseline) magnitude pruning method is not trained well (alpha $<$ 2.0). The ESD plots of the three types of training (dense, LEGCNet-PT, LEGCNet-FT) manifest a heavy-tailed distribution of eigenvalues on each layer indicating the layers are well-trained (figure \ref{fig:WW1}). A careful observation at figure \ref{fig:WW1}   reveals the following: ESD plot of a layer, where the orange spike on the far right is the tell-tale clue; it's called a Correlation Trap, A Correlation Trap refers to a situation where the empirical spectral distributions (ESDs) of the actual (green) and random (red) distributions appear remarkably similar, with the exception of a small correlation shelf located just to the right of 0. In the random ESD (red), the largest eigenvalue (orange) is noticeably positioned further to the right and is separated from the majority of the ESD's bulk. This phenomenon indicates the presence of strong correlations in the layer, which can potentially affect the overall behavior and performance of the network. Layers have an overlap of random and original ones when they have not been trained properly because they look almost random, with only a little bit of information present. And the information the layer learned may even be spurious. This is the case of a well-trained layer.

\newpage

\section{Remaining WW plots for random and magnitude pruning}

\begin{figure}[h]
\centering
\includegraphics[width=.33\textwidth]{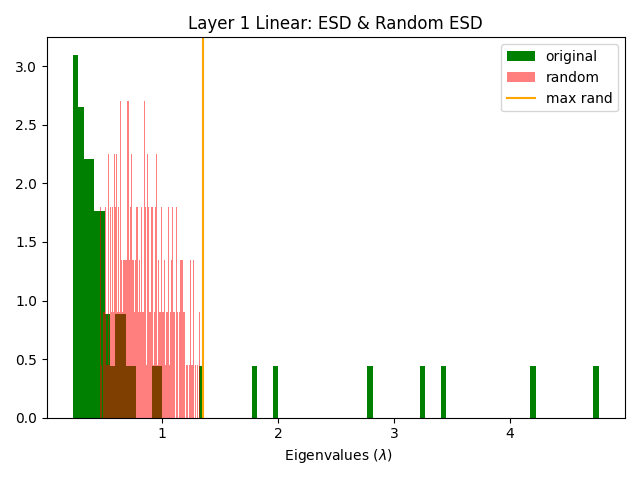}
\includegraphics[width=.33\textwidth]{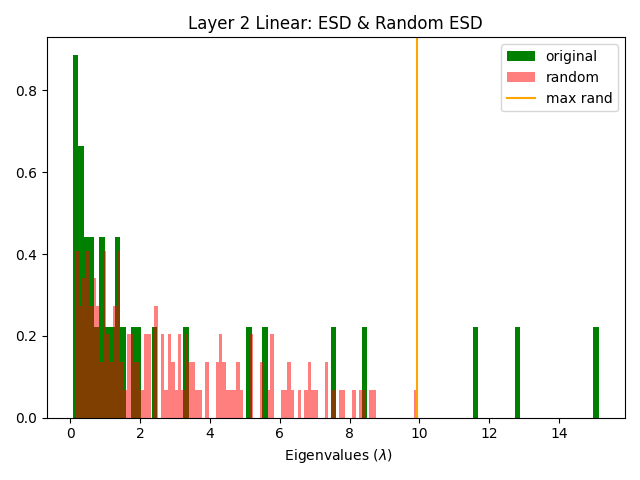} 
\includegraphics[width=.33\textwidth]{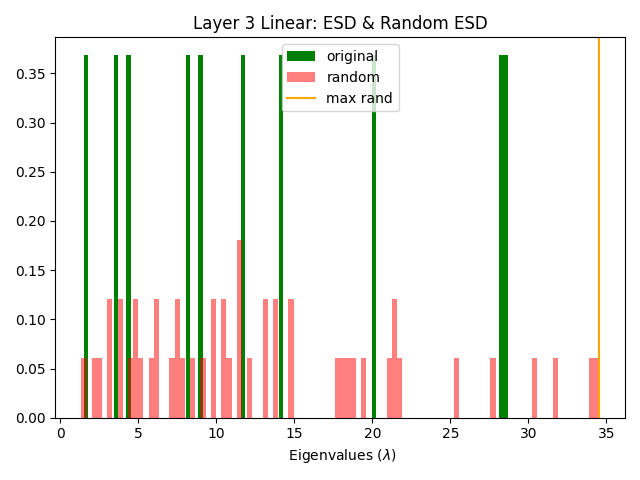} 

\caption{WW plots for Random-Pruned network}
\label{fig:figure3}
\end{figure}

\begin{figure}[h]
\centering
\includegraphics[width=.33\textwidth]{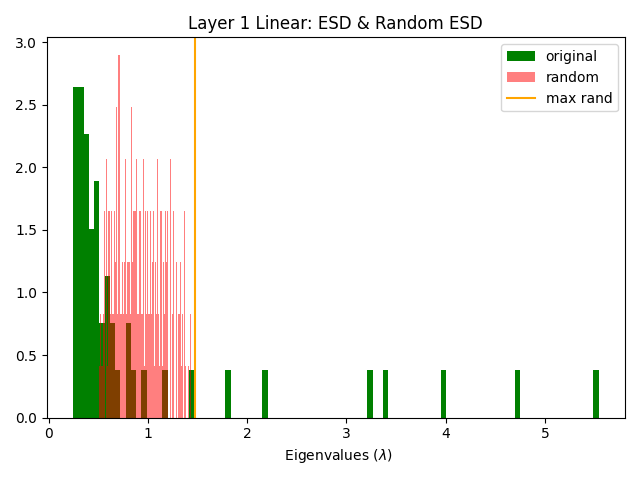} 
\includegraphics[width=.33\textwidth]{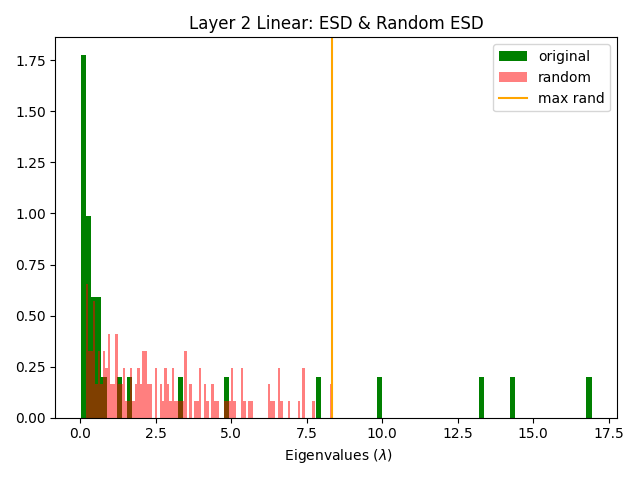} 
\includegraphics[width=.33\textwidth]{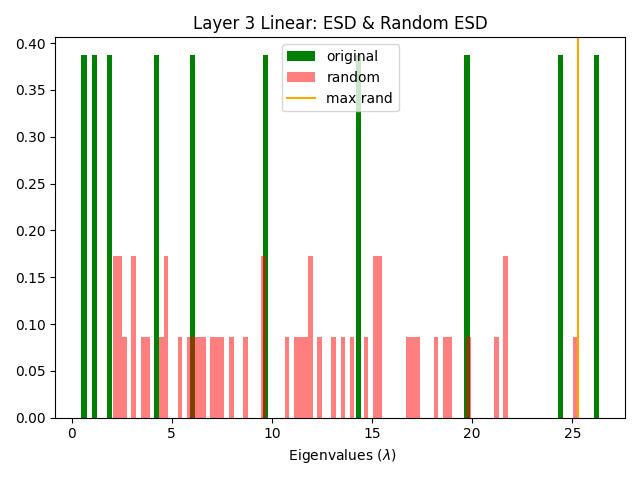} 

\caption{WW plots for Magnitude-Pruned network}
\label{fig:figure4}
\end{figure}
\newpage 

\section{Shap values and feature importance computed on Vowel, Banknote, and Titanic datasets for the 2 models - Random pruning Network, Magnitude based pruning network}

\begin{figure}[h]

\centering
\includegraphics[width=.45\textwidth]{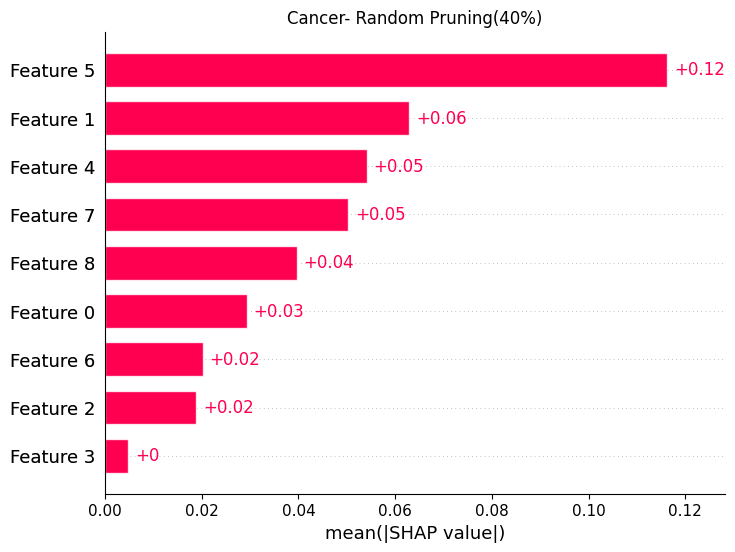} 
\includegraphics[width=.45\textwidth]{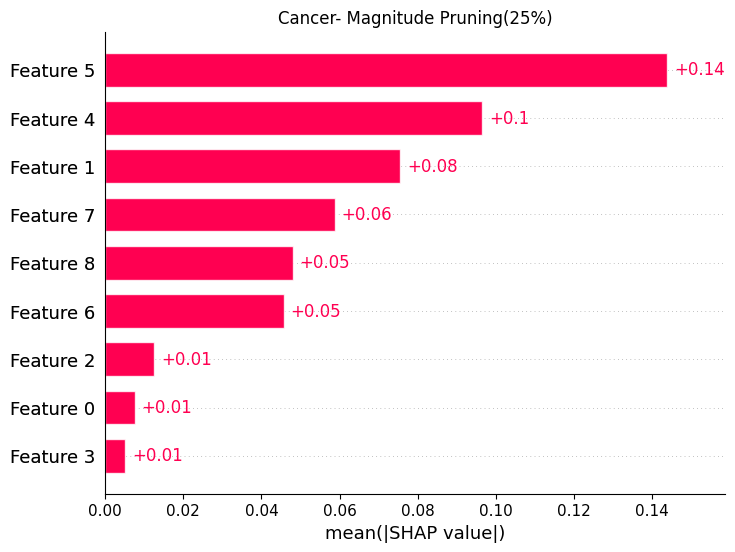} 

\includegraphics[width=.45\textwidth]{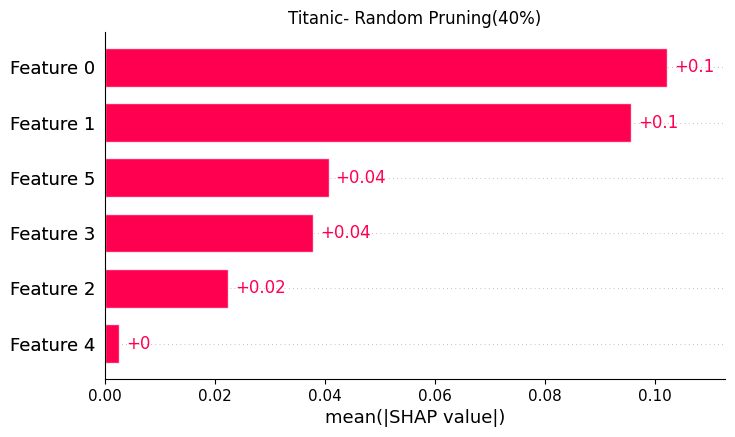} 
\includegraphics[width=.45\textwidth]{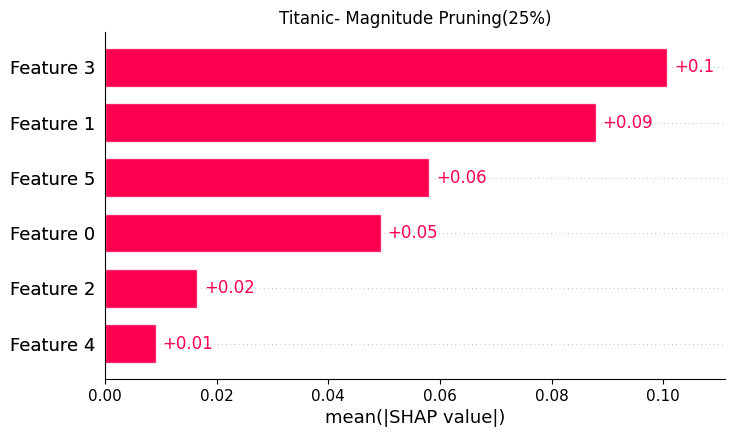} 

\includegraphics[width=.45\textwidth]{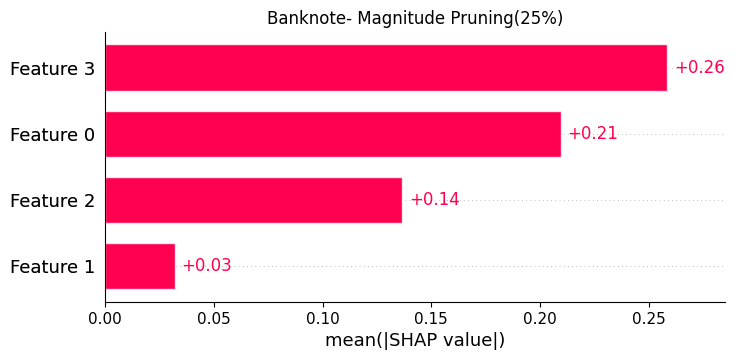}
\includegraphics[width=.45\textwidth]{media/banknote-magnitude1.png}
\caption{Shap values and feature importance computed on Vowel, Banknote, and Titanic datasets for the 2 models - Random pruning Network, Magnitude based pruning network; the feature importance for the dense network is different from the original dense network.}
\label{fig:figure5}
\end{figure}

\newpage
\section{Shap values and feature importance computed on Cancer, Banknote and Titanic datasets for all the three models - Dense Network, LEGCNet-FT and LEGCNet-PT}

\begin{figure*}[h]
\centering
\includegraphics[width=.33\textwidth]{media/cancer_dn.png}
\includegraphics[width=.33\textwidth]{media/cancer-legcnet-ft.png}
\includegraphics[width=.33\textwidth]{media/cancer-legcnet-pt.png}
\includegraphics[width=.33\textwidth]{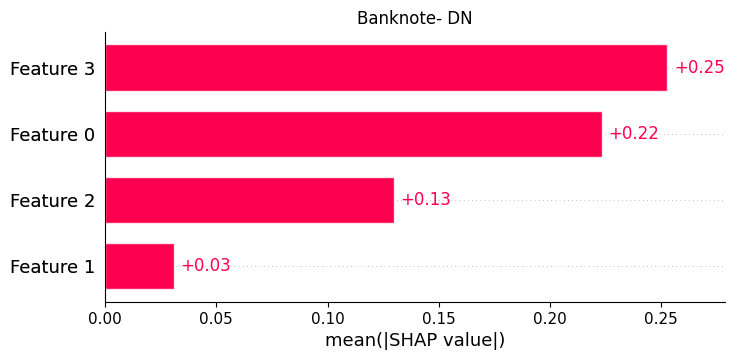}
\includegraphics[width=.33\textwidth]{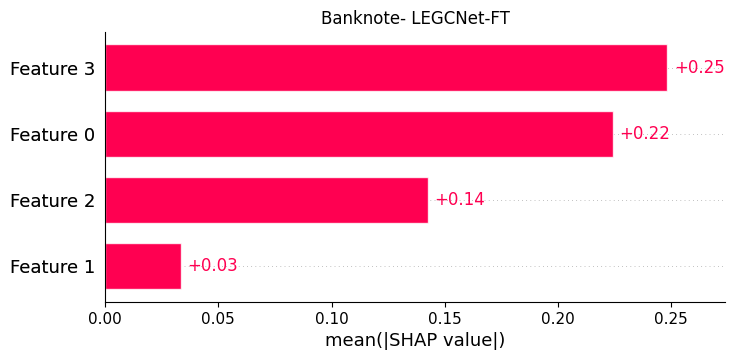}
\includegraphics[width=.33\textwidth]{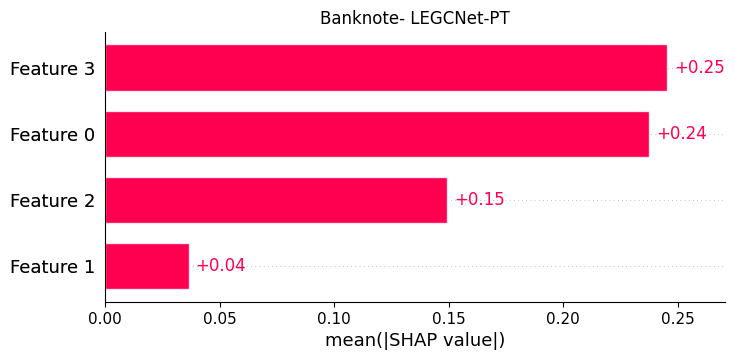}
\includegraphics[width=.33\textwidth]{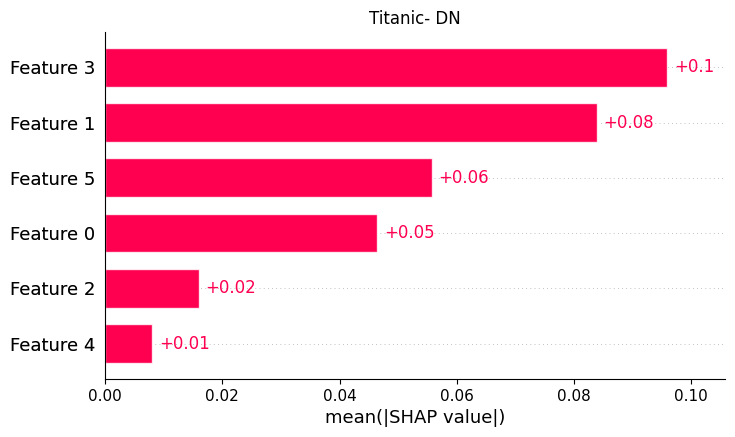}
\includegraphics[width=.33\textwidth]{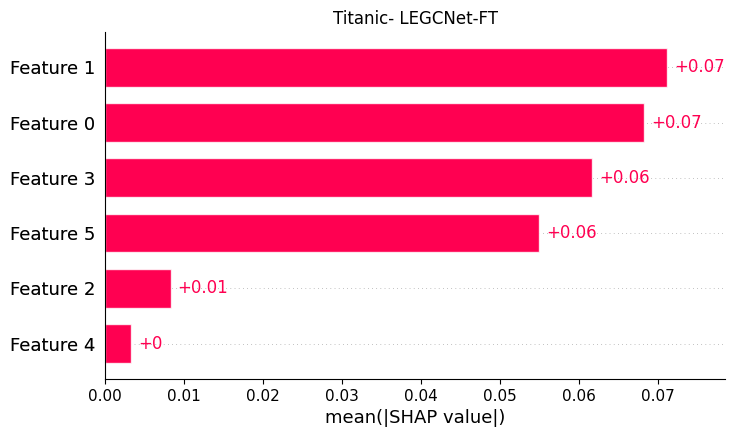}
\includegraphics[width=.33\textwidth]{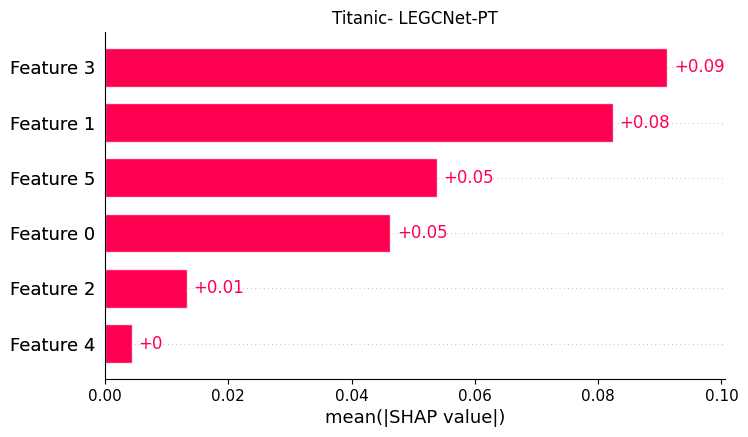}
\caption{Shap values and feature importance computed on Cancer, Banknote and Titanic datasets for all the three models - Dense Network, LEGCNet-FT and LEGCNet-PT; the feature importance for the dense network is same as LEGCNet-FT and LEGCNet-PT }
\label{fig:shap1}
\end{figure*}

\end{document}